\documentclass[letterpaper]{article} 
\usepackage{arxiv}  
\usepackage{times}  
\usepackage{helvet}  
\usepackage{courier}  
\usepackage[hyphens]{url}  
\usepackage{graphicx} 
\usepackage{natbib}  
\usepackage{caption} 
\usepackage{algorithm}
\usepackage{algorithmic}
\usepackage{multirow} 
\usepackage{booktabs} 
\usepackage{colortbl}
\usepackage{xcolor} 
\usepackage{amsmath} 
\usepackage{amssymb} 
\usepackage{amsthm} 
\newtheorem{theorem}{Theorem}
\usepackage{bbold}
\usepackage{subcaption}
  
\definecolor{green}{rgb}{0.0, 0.5, 0.0}
\definecolor{red}{rgb}{0.5, 0.0, 0.0}
\definecolor{lightgray}{gray}{0.9}

\newcommand{\greendown}{\textcolor{green}{$\blacktriangledown$}}
\newcommand{\redup}{\textcolor{red}{$\blacktriangle$}}

\newcommand{\cellgray}{\cellcolor{lightgray}}
\usepackage{newfloat}
\usepackage{listings}
\DeclareCaptionStyle{ruled}{labelfont=normalfont,labelsep=colon,strut=off} 
\floatstyle{ruled}
\newfloat{listing}{tb}{lst}{}
\floatname{listing}{Listing}
\title{Feature Clipping for Uncertainty Calibration}

\author {
    Linwei Tao\textsuperscript{\rm 1},
    Minjing Dong\textsuperscript{\rm 2},
    Chang Xu\textsuperscript{\rm 1}
}
\affiliations {
    \textsuperscript{\rm 1}University of Sydney\\
    \textsuperscript{\rm 2}City University of Hong Kong\\
    linwei.tao@sydney.edu.au, minjdong@cityu.edu.hk, c.xu@sydney.edu.au
}

\begin{document}

\maketitle

\begin{abstract}
Deep neural networks (DNNs) have achieved significant success across various tasks, but ensuring reliable uncertainty estimates, known as model calibration, is crucial for their safe and effective deployment. Modern DNNs often suffer from overconfidence, leading to miscalibration. We propose a novel post-hoc calibration method called feature clipping (FC) to address this issue. FC involves clipping feature values to a specified threshold, effectively increasing entropy in high calibration error samples while maintaining the information in low calibration error samples. This process reduces the overconfidence in predictions, improving the overall calibration of the model. Our extensive experiments on datasets such as CIFAR-10, CIFAR-100, and ImageNet, and models including CNNs and transformers, demonstrate that FC consistently enhances calibration performance. Additionally, we provide a theoretical analysis that validates the effectiveness of our method. As the first calibration technique based on feature modification, feature clipping offers a novel approach to improving model calibration, showing significant improvements over both post-hoc and train-time calibration methods and pioneering a new avenue for feature-based model calibration.
\end{abstract}

\begin{figure*}
    \centering
    \begin{subfigure}{0.68\textwidth}
        \centering
        \includegraphics[width=\linewidth]{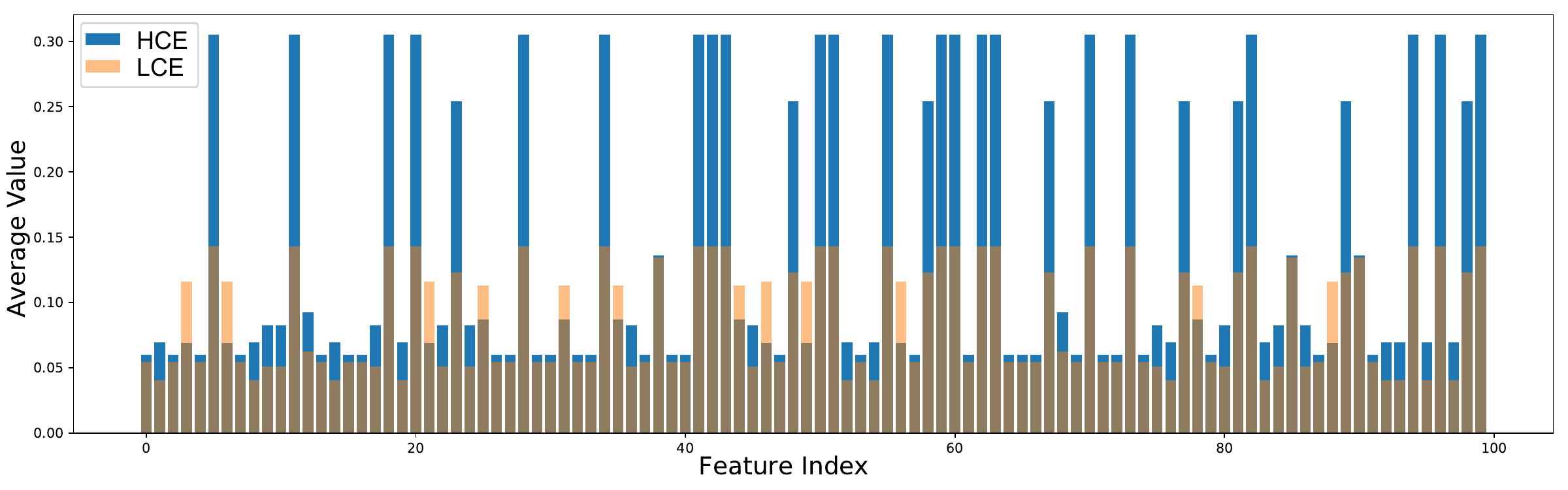}
        \caption{\textbf{Average feature value of samples with high or low calibration error.}~We randomly select 100 feature units out of 2048 units. The high/low calibration error samples are selected as the wrongly/correctly predicted samples with confidence larger than 0.95. High calibration error samples shows a obvious tendency of higher feature value in around 30\% feature. We provide a comparison of full 2048 feature units in Figure~\ref{fig:full_feature_distribution} in Appendix, which shows similar pattern.}
        \label{fig:random_feature_distribution}
    \end{subfigure}
    \hfill
    \begin{subfigure}{0.3\textwidth}
        \centering
    \includegraphics[width=\linewidth]{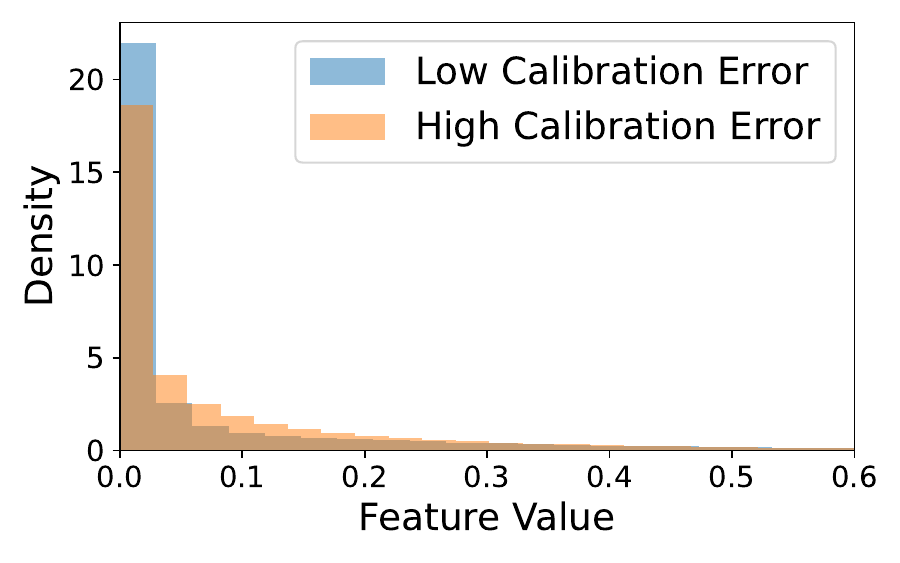}
    \caption{\textbf{Histogram of feature values for HCE and LCE samples.} The thicker tail of the HCE distribution indicates a larger variance $\sigma^2$ compared to the LCE distribution. These experiments were conducted using ResNet-50 on the CIFAR-10 dataset.}
    \label{fig:feature value histogram density}
    \end{subfigure}
    \caption{Feature Difference Between High Calibration Error Samples and Low Calibration Error samples.}
    \label{fig:mainfig}
\end{figure*}
\section{Introduction}
While deep neural networks achieve significant improvements across various tasks, model calibration—ensuring a model provides reliable uncertainty estimates—is as important as achieving high prediction accuracy. Accurate and reliable uncertainty estimation is vital for many safety-critical downstream tasks, such as autonomous driving~\cite{feng2019can} and medical diagnosis~\cite{chen2018calibration}. However, recent studies~\cite{guo2017calibration} have found that most modern neural networks struggle to accurately reflect the actual probabilities of their predictions through their confidence scores. Thus, improving model calibration techniques is essential to enhance the reliability of these models. 

Efforts to address this issue can be divided into two streams: train-time calibration and post-hoc calibration. The first stream is train-time calibration, which includes training frameworks~\cite{tao2023calibrating}, data augmentation~\cite{wang2023pitfall, zhang2022and, hendrycks2019augmix}, and regularization techniques like label smoothing~\cite{muller2019does} and entropy regularizers~\cite{pereyra2017regularizing}. Training losses such as dual focal loss~\cite{tao2023dual} and focal loss~\cite{mukhoti2020calibrating} are also notable methods. The second stream is post-hoc calibration methods, which are applied to trained models and modify the output probability. Representative works include Isotonic Regression~\cite{zadrozny2002transforming}, Histogram Binning~\cite{zadrozny2001obtaining}, and Temperature Scaling (TS)~\cite{guo2017calibration}. 
Among these, TS is widely accepted due to its simplicity and good performance. Many subsequent works~\citep{frenkel2021network, xiong2023proximity, yang2024beyond, tomani2022parameterized} propose improved versions of TS, often making the temperature factor adaptive according to different criteria.

\citet{guo2017calibration} identified overconfidence as a major cause of miscalibration in most modern neural networks. Adding a maximum-entropy penalty effectively increases prediction uncertainty, thereby mitigating overconfidence issues. Many calibration methods can be summarized as using entropy regularization in various forms. \citet{pereyra2017regularizing} apply a maximum-entropy penalty uniformly to all samples. Similarly, Label Smoothing~\citep{muller2019does} can be transformed into a form of entropy penalty, while Focal Loss~\citep{mukhoti2020calibrating} can be viewed as the upper bound of a form with negative entropy, effectively adding a maximum-entropy regularizer. TS often uses a temperature parameter larger than 1, resulting in a smoother probability distribution with higher entropy.

Since the features extracted by neural networks are direct representations of data, a possible way to mimic this maximum-entropy penalty effect is by applying information loss directly to the features, thereby increasing entropy. To explore this idea, we begin by comparing high calibration error samples with low calibration error samples. Accurately obtaining per-sample calibration error is non-trivial, so we choose wrongly predicted samples with high confidence (greater than 0.95) as high calibration error (HCE) samples and correctly predicted samples with high confidence (greater than 0.95) as low calibration error (LCE) samples. We randomly select 100 feature units from the feature of these samples and plot the average unit value in Figure~\ref{fig:random_feature_distribution}. We observe that the feature value of HCE samples is much higher than that of LCE samples in some units. A potential solution is to clip the feature values, making values larger than a threshold $c$ equal to $c$. This might help reduce the abnormally large feature values, increasing the entropy of HCE samples. For example, in Figure~\ref{fig:random_feature_distribution}, we propose feature clipping at 0.15 to increase the entropy of HCE samples while retaining the information of LCE samples. This removes significant information from HCE samples, making them more uncertain, while maintaining as much information as possible from LCE samples. We also plot the histogram of feature of both HCE samples and LCE samples to examine the feature distribution of both samples as shown in Fig.~\ref{fig:feature value histogram density}. We observe that HCE samples exhibit a thicker tail, indicating a larger variance compared to LCE samples. These patterns suggest notable differences in features between HCE and LCE samples. Therefore, it is worthwhile to conduct a deeper study on how to calibrate models based on these features.

Motivated by these observation, we propose a simple and effective post-hoc calibration method called \textit{feature clipping} (FC), which clips the feature value to a hyperparameter $c$, optimized on the validation set to minimize negative log likelihood (NLL), similar to temperature scaling. We also provide a solid theoretical analysis to prove the effectiveness of FC. To the best of our knowledge, we are the first to propose a calibration method based on feature modification. We conduct extensive experiments on a wide range of datasets, including CIFAR-10, CIFAR-100, and ImageNet, and models, including CNNs and transformers. Our method shows consistent improvement. Furthermore, since we are the first to perform calibration on features, our method is orthogonal to previous calibration methods. Extensive experiments demonstrate that FC can enhance calibration performance over both previous post-hoc and train-time calibration methods. Overall, we make the following contributions:

\begin{itemize}
    \item We propose a simple and effective calibration method called feature clipping, which achieves SOTA calibration performance across multiple models and datasets.
    \item We provide a solid theoretical analysis to prove the effectiveness of feature clipping by showing feature clipping increases more entropy on HCE samples.
    \item We are the first to propose calibration based on features, initiating a new avenue for feature-based calibration. Our method serves as a strong baseline for this emerging area.
\end{itemize}

\section{Related Works}

Deep neural networks have long been a focus of calibration research~\citep{guo2017calibration}, with extensive studies examining their calibration properties~\citep{minderer2021revisiting, wang2021rethinking, tao2023benchmark}. Numerous calibration methods have been proposed, generally divided into two categories: train-time calibration and post-hoc calibration.

\subsubsection{Train-Time Calibration}Train-time calibration aims to improve a model's calibration performance during training. A notable example is focal loss~\citep{mukhoti2020calibrating}, with subsequent works such as dual focal loss~\citep{tao2023dual} focusing on both the highest and second-highest probabilities. Adaptive focal loss~\citep{ghosh2022adafocal} modifies hyperparameters for different sample groups based on prior training knowledge. These focal loss-based methods can be transformed into an upper bound of negative entropy, thereby performing an entropy penalty during training. Similarly, label smoothing~\citep{muller2019does} can also be transformed into a form of entropy penalty.

\subsubsection{Post-Hoc Calibration}
Post-hoc calibration is resource-efficient and can be easily applied to pretrained models without altering their weights, preserving the model's accuracy and robustness. A common technique is temperature scaling (TS), which adjusts the output probability distribution's sharpness via a temperature parameter optimized to minimize negative log likelihood (NLL) on a validation set. TS typically uses larger temperature parameters for CNN models, reducing probability distribution sharpness and acting as an uniform maximum-entropy regularizer. Many subsequent methods aim to improve TS by applying adaptive temperature parameters, treating samples differently for a more effective maximum-entropy regularizer. For example, CTS~\citep{frenkel2021network} adapts temperature based on class labels, while PTS~\citep{tomani2022parameterized} proposes learnable temperature parameters using a neural network. Recent methods like Proximity-based TS~\citep{xiong2023proximity} and Group Calibration~\citep{yang2024beyond} adjust temperature based on features, aiming for more precise entropy penalties.

\subsubsection{Calibration Using Features}
Although feature representation is a crucial aspect of deep neural networks and is well-studied in robustness literature~\citep{ilyas2019adversarial}, it is underutilized in calibration literature. Pioneering works such as~\cite{xiong2023proximity, yang2024beyond} have explored using features to group similar samples to achieve multi-calibration~\citep{hebert2018multicalibration}. However, they do not perform calibration based on feature modification.

\section{Methodology}
\subsubsection{Problem Formulation}

In a classification task, let \(\mathcal{X}\) be the input space and \(\mathcal{Y}\) be the label space. The classifier \(f\) maps an input to a probability distribution \(\hat{p}_{[1,2,\ldots,K]} \in [0,1]^K\) over \(K\) classes. The confidence of a prediction is defined as the largest probability, \(\max(\hat{p}_i)\). For simplicity, we use \(\hat{p}\) to represent confidence in the following discussion.

A network is perfectly calibrated if the predicted confidence \(\hat{p}\) accurately represents the true probability of the classification being correct. Formally, a perfectly calibrated network satisfies \(\mathbb{P}(\hat{y}=y|\hat{p}=p) = p\) for all \(p \in [0,1]\)~\citep{guo2017calibration}, where \(\hat{y}\) is the predicted label and \(y\) is the ground truth label. Given the confidence score and the probability of correctness, the \textit{Expected Calibration Error} (ECE) is defined as \(\mathbb{E}_{\hat{p}}[|\mathbb{P}(\hat{y}=y|\hat{p})-\hat{p}|]\).
In practice, since the calibration error cannot be exactly derived from finite samples, an approximation of ECE is introduced~\citep{guo2017calibration}. Specifically, samples are grouped into \(M\) bins \(\{B_m\}_{m=1}^M\) based on their confidence scores, where \(B_m\) contains samples with confidence scores \(\hat{p}_i \in \left[\frac{m-1}{M},\frac{m}{M}\right)\). For each bin \(B_m\), the average confidence is computed as 
\(C_m = \frac{1}{|B_m|} \sum_{i \in B_m} \hat{p}_i\)
and the bin accuracy as \(A_m = \frac{1}{|B_m|} \sum_{i \in B_m} \mathbb{1}(\hat{y}_i = y_i),\)
where \(\mathbb{1}\) is the indicator function. The ECE is then approximated as the expected absolute difference between bin accuracy and average confidence:

\[ \text{ECE} \approx \sum_{m=1}^M \frac{|B_m|}{N} \left| A_m - C_m \right|,\]
where \(N\) is the total number of samples. Besides this estimated ECE, there are variants like Adaptive ECE~\citep{krishnan2020improving}, which groups samples into bins with equal sample sizes, and Classwise ECE~\citep{kull2019beyond}, which computes ECE over \(K\) classes.

\subsubsection{Feature Clipping}

We propose feature clipping (FC), a simple and effective post-hoc calibration method designed to reduce overconfidence problem in deep neural networks. The key idea is to clip the feature values to a specified threshold, thereby increasing entropy in HCE samples while preserving the information in LCE samples. This approach helps mitigate overconfidence issues in HCE samples and improves overall model calibration. Given feature values \(x\), we apply feature clipping as follows:

\begin{equation}
    \tilde{x} = \min(\max(x, c), -c)
    \label{eq:feature_clipping}
\end{equation}
where \(c\) is a positive hyperparameter optimized on a validation set to minimize negative log likelihood (NLL).

\section{Theoretical Evidence}

In this section, we present theoretical evidence to explain the effectiveness of feature clipping by analyzing the information loss in features of HCE and LCE samples. Our aim is to demonstrate that after feature clipping, HCE samples, characterized by larger variance, experience greater information loss compared to LCE samples, which have smaller variance. Consequently, we perform the entropy penalty differently to HCE and LCE samples and make HCE samples more uncertain.

\subsubsection{Entropy of original feature}We consider the case where the output are all postive values after ReLU activation function. Consider the feature vector $ \mathbf{x} := \{x_1, \ldots, x_n\} $ extracted from a sample, which is normally the output of penultimate layer of a neural network. Suppose the feature value $X$ follows a rectified normal distribution~\citep{socci1997rectified}, which is a mixture distribution with both discrete variables and continuous variables. To calculate the entropy for this mixture distribution\footnote{For a mixture distribution that includes both discrete and continuous variables, we treat the entropy of the discrete and continuous parts individually. This means the overall entropy is a weighted combination of the entropy of these two parts.}~\citep{politis1991entropy}, first, we treat the continuous variables as the truncated normal distribution~\cite{burkardt2014truncated}.

For a standard truncated normal distribution, suppose $X$ has a normal distribution with mean $\mu=0$ and variance $\sigma^2$ and lies within the interval $(a,b)$. The probability density function (PDF) of truncated normal distribution is given by:
\begin{equation}
f(x; \mu, \sigma, a, b) = \frac{1}{\sigma} \frac{\phi\left(\frac{x - \mu}{\sigma}\right)}{\Phi\left(\frac{b - \mu}{\sigma}\right) - \Phi\left(\frac{a - \mu}{\sigma}\right)}
\end{equation}
and by $f = 0$ otherwise.
Here,
$\phi(\xi) = \frac{1}{\sqrt{2\pi}} \exp\left(-\frac{1}{2} \xi^2 \right)$
is the probability density function of the standard normal distribution and $\Phi(\cdot)$ is its cumulative distribution function $\Phi(x) = \frac{1}{2} \left(1 + \text{erf}\left(\frac{x}{\sqrt{2}}\right)\right)$ and $\text{erf}(x) = \frac{2}{\sqrt{\pi}} \int_{0}^{x} e^{-t^2} \, dt$ is the error function. 
By definition, if $b = \infty$, then $\Phi\left(\frac{b - \mu}{\sigma}\right) = 1$. 
The entropy of truncated normal distribution is given by: 
\begin{align}
    H_c(x; \mu, \sigma, a, b) &= - \int_a^b f(x) \log f(x) \, dx \notag\\
    &=\log(\sqrt{2 \pi e \sigma Z}) + \frac{\alpha \phi(\alpha) - \beta \phi(\beta)}{2Z}
\end{align}
where $\alpha = \frac{a - \mu}{\sigma}, \quad \beta = \frac{b - \mu}{\sigma}$ and $Z = \Phi(\beta) - \Phi(\alpha)$. 

Thus, the PDF of the continuous variables of the mixture distribution is $f(x; 0, \sigma, 0, +\infty)$ and the corresponding differential entropy is $H_c(x; 0, \sigma, 0, +\infty)$. The probability mass function (PMF) for discrete variables in the mixture distribution is given by:
\begin{equation}
        p(x) = 
\begin{cases} 
100\% & \text{if } x=0\\
0 & \text{otherwise}\\
\end{cases},
\end{equation}
and the corresponding Shannon entropy is $H_d(x) = 0$.

Assume the input of ReLU layer follows Gaussian distribution with mean at 0, we can derive that feature $x$ with probability $q=0.5$ to be discrete variables and $1-q$ to be continuous variables. According to the entropy calculation of mixture distribution ~\citep{politis1991entropy}, the entropy of original feature $x$ is given by:
\begin{align}
    H(x) = &-q \log q - (1 - q) \log (1 - q)\notag\\
    & + qH_d(x)+ (1-q)H_c(x; 0, \sigma, 0, +\infty)\notag\\
    = &-\log(\frac{1}{2})-\frac{1}{2}H_c(x; 0, \sigma, 0, +\infty)\notag\\
    = &-\log(\frac{1}{2})-\frac{1}{2}\log(\sqrt{\pi e \sigma})
\end{align}

\subsubsection{Entropy of clipped feature} Similarly, the clipped feature $\tilde{x}$ follows mixture distribution with discrete variables and continuous variables, where the PDF of the continuous variables is $f(\tilde{x}; 0, \sigma, 0, c)$ and the corresponding differential entropy is $H_c(x; 0, \sigma, 0, c)$ The PMF for discrete variables is given by:
\begin{equation}
        p(\tilde{x}) = 
\begin{cases} 
\frac{\Phi(0)}{\Phi(0)+(1-\Phi\left(\frac{c}{\sigma}\right))} & \text{if } \tilde{x}=0\\
\frac{1-\Phi\left(\frac{c}{\sigma}\right)}{\Phi(0)+(1-\Phi\left(\frac{c}{\sigma}\right))} & \text{if } \tilde{x}=c\\
0 & \text{otherwise}\\
\end{cases},
\end{equation}
and the corresponding entropy is 
\begin{align}
    H_d(\tilde{x})=&-\sum p(\tilde{x})\log(p(\tilde{x})) \notag\\
    =&-\frac{0.5}{1.5-\Phi\left(\frac{c}{\sigma}\right)}\log\left(\frac{0.5}{1.5-\Phi\left(\frac{c}{\sigma}\right)}\right)\notag\\
    &-\frac{1-\Phi\left(\frac{c}{\sigma}\right)}{1.5-\Phi\left(\frac{c}{\sigma}\right)}\log\left(\frac{1-\Phi\left(\frac{c}{\sigma}\right)}{1.5-\Phi\left(\frac{c}{\sigma}\right)}\right)
\end{align}
Since $\tilde{x}$ with probability $\tilde{q}=\Phi(0)+(1-\Phi\left(\frac{c}{\sigma}\right))$ to be discrete variables and $1-\tilde{q}$ to be continuous variables, similarly, the entropy of clipped feature $\tilde{x}$ can be derived as following form according to~\citep{politis1991entropy},
\begin{align}
    H(\tilde{x}) = &-\tilde{q} \log \tilde{q} - (1 - \tilde{q}) \log (1 - \tilde{q})\notag\\
    & + \tilde{q}H_d(\tilde{x})+ (1-\tilde{q})H_c(\tilde{x}; 0, \sigma, 0, c)\notag\\
    = &-\left(1.5-\Phi\left(\frac{c}{\sigma}\right)\right)\log\left(1.5-\Phi\left(\frac{c}{\sigma}\right)\right)\notag\\
    &-\left(\Phi\left(\frac{c}{\sigma}\right)-0.5\right)\log\left(\Phi\left(\frac{c}{\sigma}\right)-0.5\right)\notag\\
    &-0.5\log\left(\frac{0.5}{1.5-\Phi\left(\frac{c}{\sigma}\right)}\right)\notag\\
    &-\left(1-\Phi\left(\frac{c}{\sigma}\right)\right)\log\left(\frac{1-\Phi\left(\frac{c}{\sigma}\right)}{1.5-\Phi\left(\frac{c}{\sigma}\right)}\right)\notag\\
    & + (\Phi\left(\frac{c}{\sigma}\right)-0.5)H_c(\tilde{x}; 0, \sigma, 0, c)
\end{align}

\begin{table}[!ht]
\centering
\begin{tabular}[width=\linewidth]{c|ccc}
\toprule
 & \textbf{$H_{\text{sm}}(X)$}  & \textbf{$H_{\text{sm}}(\tilde{X})$} & \textbf{$\Delta H_{\text{sm}}$}\\ \midrule
HCE & 0.0824& 0.5723& \textbf{0.4908}\\
LCE & 0.0032& 0.1525& 0.1493\\ 
\bottomrule
\end{tabular}
\caption{Entropy values calculated on Softmax probability before and after clipping, and their differences for HCE and LCE samples. FC makes HCE samples more uncertain. The experiment is conducted on ResNet-50 on CIFAR-10.}
\label{table:softmax entropy difference}
\end{table}

\subsubsection{Entropy Difference}Then, the Shannon entropy difference between features before and after clipping is given by
\begin{align}
    \Delta H =&-\left(\Phi\left(\frac{c}{\sigma}\right)-0.5\right)\log\left(\Phi\left(\frac{c}{\sigma}\right)-0.5\right)\notag\\
&+0.5\log\left(0.5\right)\notag\\
    &-\left(1-\Phi\left(\frac{c}{\sigma}\right)\right)\log\left(1-\Phi\left(\frac{c}{\sigma}\right)\right)\notag\\
    & + (\Phi\left(\frac{c}{\sigma}\right)-0.5)H_c(\tilde{x}; 0, \sigma, 0, c)\notag\\
    &+\frac{1}{2}\log(\sqrt{\pi e \sigma})
\end{align}
and $\Delta H$ is determined by the clipping threshold $c$ and $\sigma$. We adopt the empirical result that $\sigma_{HCE} > \sigma_{LCE}$, as discussed in previous section. Thus, we can derive the theorem,

\begin{theorem}
High calibration error samples suffer larger entropy difference compared to low calibration error samples after feature clipping.\[\Delta H_{LCE}<\Delta H_{HCE}\]
\label{theorem:entropy difference}
\end{theorem}

\begin{table*}[!ht]
	\centering
	\scriptsize
	\resizebox{\linewidth}{!}{%
		\begin{tabular}{ccccccccccccccc}
			\toprule
			\textbf{Dataset} & \textbf{Model} & \multicolumn{2}{c}{\textbf{Original Feature}} &\multicolumn{2}{c}{\textbf{TS}} &\multicolumn{2}{c}{\textbf{ETS}} &
			\multicolumn{2}{c}{\textbf{PTS}} &
   \multicolumn{2}{c}{\textbf{CTS}} &
   \multicolumn{2}{c}{\textbf{Group Calibration}} & 
            \\
			\textbf{} & \textbf{} & \multicolumn{2}{c}{ } &
			\multicolumn{2}{c}{\citep{guo2017calibration}} &
   \multicolumn{2}{c}{\citep{zhang2020mix}} &
   \multicolumn{2}{c}{\citep{tomani2022parameterized}} &
    \multicolumn{2}{c}{\citep{frenkel2021network}} &
    \multicolumn{2}{c}{\citep{yang2024beyond}} &

           \\
			&& base & +ours(c) & base & +ours & base & +ours & base & +ours & base & 
			+ours & base & +ours \\
			\midrule
			\multirow{3}{*}{CIFAR-10} 
            & ResNet-50&4.34&\cellgray1.10(0.23) \greendown&1.39&\cellgray1.22 \greendown&1.37&\cellgray1.22 \greendown&1.36&\cellgray1.25 \greendown&1.46&\cellgray1.25 \greendown&0.97&\cellgray\textbf{0.49} \greendown\\
			& ResNet-110&4.41&\cellgray0.96(0.23) \greendown&0.98&\cellgray0.94 \greendown&0.98&\cellgray0.94 \greendown&0.95&\cellgray\textbf{0.90} \greendown&1.13&\cellgray\textbf{0.90} \greendown&1.24&\cellgray1.78 \redup\\
            & DenseNet-121&4.51&\cellgray\textbf{1.05}(0.45) \greendown&1.41&\cellgray1.11 \greendown&1.40&\cellgray1.12 \greendown&1.38&\cellgray1.14 \greendown&1.44&\cellgray1.14 \greendown&1.27&\cellgray2.51 \redup\\
			\midrule
			\multirow{3}{*}{CIFAR-100}
            & ResNet-50&17.52&\cellgray3.98(0.60) \greendown&5.72&\cellgray4.26 \greendown&5.68&\cellgray4.29 \greendown&5.64&\cellgray4.37 \greendown&6.03&\cellgray4.37 \greendown&3.43&\cellgray\textbf{1.70} \greendown\\
			& ResNet-110&19.06&\cellgray4.40(0.61) \greendown&5.12&\cellgray4.81 \greendown&5.10&\cellgray4.81 \greendown&5.05&\cellgray4.98 \greendown&5.43&\cellgray4.98 \greendown&\textbf{2.71}&\cellgray3.45 \redup\\
			& DenseNet-121&20.99&\cellgray3.28(1.40) \greendown&5.15&\cellgray3.92 \greendown&5.09&\cellgray3.95 \greendown&5.06&\cellgray4.04 \greendown&4.87&\cellgray4.04 \greendown&2.84&\cellgray\textbf{1.75} \greendown\\
			\midrule
			\multirow{4}{*}{ImageNet}& ResNet-50&3.69&\cellgray1.74(2.06) \greendown&2.08&\cellgray1.64 \greendown&2.08&\cellgray1.65 \greendown&2.11&\cellgray1.63 \greendown&3.05&\cellgray1.63 \greendown&1.30&\cellgray\textbf{1.00} \greendown\\
            & DenseNet-121&6.66&\cellgray3.08(3.45) \greendown&1.65&\cellgray1.19 \greendown&1.65&\cellgray1.20 \greendown&1.61&\cellgray1.20 \greendown&2.21&\cellgray1.20 \greendown&2.67&\cellgray\textbf{0.63} \greendown\\
            & Wide-Resnet-50&5.52&\cellgray2.52(3.05) \greendown&3.01&\cellgray2.21 \greendown&3.01&\cellgray2.20 \greendown&3.00&\cellgray2.18 \greendown&4.31&\cellgray2.18 \greendown&3.01&\cellgray\textbf{0.88} \greendown\\
            & MobileNet-V2&2.72&\cellgray1.36(1.73) \greendown&1.92&\cellgray1.41 \greendown&1.92&\cellgray1.41 \greendown&1.93&\cellgray1.44 \greendown&2.34&\cellgray1.44 \greendown&1.81&\cellgray\textbf{0.50} \greendown\\
			\bottomrule
		\end{tabular}%
	}
	\caption{\textbf{ECE$\downarrow$ before and after after feature clipping.}\quad  ECE is measured as a percentage, with lower values indicating better calibration. ECE is evaluated for different post hoc calibration methods, both before (base) and after (+ours) feature clipping. The results are calculated with number of bins set as 15. The optimal $c$ is determined on the validation set, included in brackets.}
	\label{table:ece compare with posthoc methods}
\end{table*}
The detailed proof of Theorem~\ref{theorem:entropy difference} is given in Appendix. To verify our conclusion, we further calculate the entropy difference at Softmax layer, which is consistent with our observation. Specifically, we numerically calculate the entropy based on Softmax probability before and after feature clipping.
As shown in Table~\ref{table:softmax entropy difference}, both entropy of HCE samples $H_{\text{sm}}^{\text{HCE}}(X)$ and entropy of LCE samples $H_{\text{sm}}^{\text{LCE}}(X)$ are close to zero before clipping. However, after feature clipping, the entropy of HCE samples $H_{\text{sm}}^{\text{HCE}}(\tilde{X})$ become much larger than entropy of LCE samples $H_{\text{sm}}^{\text{LCE}}(\tilde{X})$. 
\[
\Delta H_{\text{sm}}^{\text{LCE}} \ll \Delta H_{\text{sm}}^{\text{HCE}}.
\]
In other words, feature clipping successfully differentiated the handling of HCE and LCE samples, increase more entropy in HCE samples compared to LCE samples and make HCE samples more uncertain.

\section{Experiments}
\subsection{Experiment Setup} 
\subsubsection{Models and Datasets}We evaluate our methods on various deep neural networks (DNNs), including ResNet~\citep{he2016deep}, Wide-ResNet~\citep{zagoruyko2016wide}, DenseNet~\citep{huang2017densely}, MobileNet~\citep{howard2017mobilenets}, and ViT~\citep{dosovitskiy2020image}, using the CIFAR-10, CIFAR-100~\citep{krizhevsky2009learning}, and ImageNet-1K~\citep{deng2009imagenet} datasets to assess the effectiveness of feature clipping. Pre-trained weights for post hoc calibration evaluation are provided by PyTorch.torchvision. Pre-trained weights trained by other train-time calibration methods are provided by ~\citet{mukhoti2020calibrating}. 

\subsubsection{Metrics}We use the Expected Calibration Error (ECE) and accuracy as our primary metrics for evaluation. Additionally, we incorporate Adaptive ECE, a variant of ECE, which groups samples into bins of equal sizes to provide a balanced evaluation of calibration performance. For both ECE and Adaptive ECE, we use bin size at 15. We also measure the influence of calibration methods on prediction accuracy. 

\subsubsection{Comparison methods}
We compare our methods with several popular and state-of-the-art (SOTA) approaches. For post-hoc methods, we evaluate the widely used temperature scaling (TS) and other subsequent methods such as ETS~\citep{zhang2020mix}, PTS~\citep{tomani2022parameterized}, CTS~\citep{frenkel2021network}, and a recently proposed SOTA calibration method called Group Calibration~\citep{yang2024beyond}. For all TS-based methods, we determine the temperature by tuning the hyperparameter on the validation set to minimize the Negative Log Likelihood (NLL). To maintain consistency with TS, we also determine the optimal clipping threshold 
$c$ on the validation set by minimizing the NLL.
For training-time calibration methods, we include training with Brier loss~\citep{brier1950verification}, MMCE loss~\citep{kumar2018trainable}, label smoothing~\citep{muller2019does} with a smoothing factor of 0.05, focal loss~\citep{mukhoti2020calibrating} with $\gamma$ set to 3, FLSD-53~\citep{mukhoti2020calibrating} using the same $\gamma$ scheduling scheme as in~\citep{mukhoti2020calibrating}, and Dual Focal Loss~\citep{tao2023dual}. Detailed settings for each method are following the settings in ~\citep{mukhoti2020calibrating}.

\subsection{Calibration Performance}
To evaluate the performance, we assess feature clipping on both post-hoc methods and train-time calibration. We also find that post-hoc calibration methods hardly improve calibration performance on ViT and provide an empirical analysis to support this finding.
\subsubsection{Compare with Post-Hoc Calibration Methods}
We compare the post-hoc calibration performance across multiple datasets and models, as shown in Table~\ref{table:ece compare with posthoc methods}. FC consistently improves over the original features. With similar computational overhead and simplicity, FC outperforms TS in most cases, as seen when comparing columns 2 and 3. When combined with other post-hoc calibration methods, FC achieves state-of-the-art results. While Group Calibration also shows competitive results, it requires training an additional neural network based on features, resulting in higher computational overhead. Additionally, feature clipping is not compatible with Group Calibration in some cases, likely because Group Calibration separates groups based on features, while FC clips features, reducing information and making them less separable. Notably, in several instances, FC alone achieves the best performance, highlighting the potential of feature-based calibration. The simplicity of FC as a baseline method suggests significant opportunities for enhancement and optimization in future work. This demonstrates that even straightforward approaches like FC can yield substantial improvements, paving the way for more sophisticated feature-based calibration techniques. We also evaluate feature clipping using Adaptive ECE, a balanced version of ECE, with the results presented in Table~\ref{table:adaptive ece compare with posthoc methods} in the Appendix. FC demonstrates competitive results in this evaluation as well.

\begin{table}
\centering
\begin{tabular}{c|ccccc}
\toprule
 &Vanilla&TS&ETS&PTS&CTS\\ 
\midrule
w/o FC & 5.24& 5.73&5.73&5.73&6.07\\ 
w/ FC &  \textbf{5.04}\greendown & 5.59\greendown&5.60\greendown&5.60\greendown&5.60\greendown\\ 
\bottomrule
\end{tabular}
\caption{\textbf{ECE Calibration performance on Vision Transformer.}\quad Feature Clipping provides little but consistent improvement on Vision Transformer. Experiments are conducted on ViT-L-16 on ImageNet.}
\label{table:ece on vit}
\end{table}

\begin{figure}
    \centering
    \includegraphics[width=\linewidth]{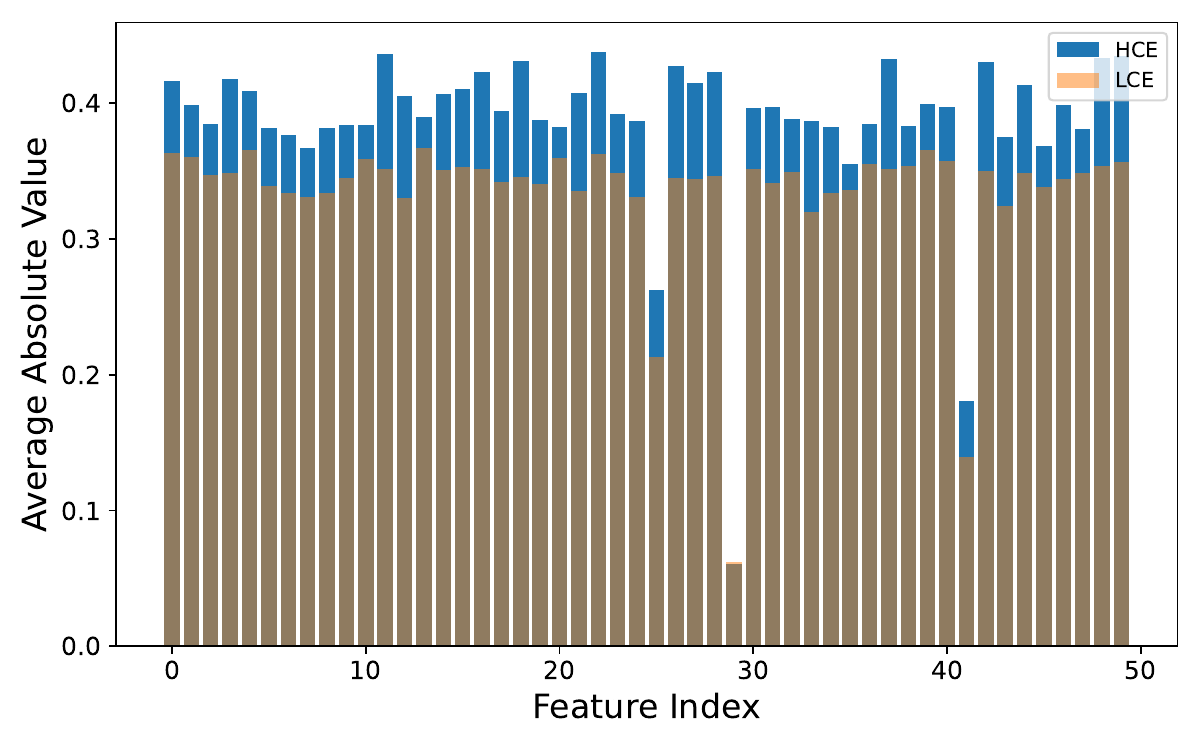}
    \caption{\textbf{Average absolute feature value of samples with high or low calibration error on Vision Transformer.}~We randomly select 50 feature units out of 2048 units. The high/low calibration error samples are selected as the wrongly/correctly predicted samples with confidence larger than 0.8.}
    \label{fig:random_feature_distribution vit}
\end{figure}

\begin{table}[!ht]
\centering
\begin{tabular}{c|cc|cc}
\toprule
 & \multicolumn{2}{c|}{ResNet-50} & \multicolumn{2}{c}{ViT-L-16} \\ 
Confidence & Correct & Wrong & Correct & Wrong \\ 
\midrule
$>0.80$ & 5921 & 471 & 6431 & 455 \\ 
$>0.90$ & 5221 & 271 & 3429 & 92 \\ 
$>0.95$ & 4526 & 161 & 44 & 0 \\ 
$>0.99$ & 3173 & 53 & 0 & 0 \\ 
\bottomrule
\end{tabular}
\caption{\textbf{Number of high confidence samples in ImageNet test set.}\quad The total number of samples is 50,000. }
\label{table:overconfidence count for r50 and vit}
\end{table}

\begin{table*}[!ht]
	\centering
	\scriptsize
	\resizebox{\linewidth}{!}{%
		\begin{tabular}{ccccccccccccccccccc}
			\toprule
			\textbf{Dataset} & \textbf{Model} &\multicolumn{2}{c}{\textbf{Cross Entropy}} &\multicolumn{2}{c}{\textbf{Brier Loss}} &\multicolumn{2}{c}{\textbf{MMCE}} &
			\multicolumn{2}{c}{\textbf{LS-0.05}} &
   \multicolumn{2}{c}{\textbf{Focal Loss}} &
   \multicolumn{2}{c}{\textbf{FLSD-53}} &
   \multicolumn{2}{c}{\textbf{Dual Focal Loss}} & 
            \\
			\textbf{} & \textbf{} &
			\multicolumn{2}{c}{} &
			\multicolumn{2}{c}{\citep{brier1950verification}} &
   \multicolumn{2}{c}{\citep{kumar2018trainable}} &
   \multicolumn{2}{c}{\citep{muller2019does}} &
    \multicolumn{2}{c}{\citep{mukhoti2020calibrating}} &
    \multicolumn{2}{c}{\citep{mukhoti2020calibrating}} &
    \multicolumn{2}{c}{\citep{tao2023dual}} &
           \\
			&& base & +ours & base & +ours & base & +ours & base & +ours & base & 
			+ours & base & +ours & base & +ours \\
			\midrule
			\multirow{4}{*}{CIFAR-10} 
            & ResNet-50
            & 4.34 &\cellgray 1.10(0.23)  \greendown
            & 1.80 &\cellgray 1.49(0.98)  \greendown
            & 4.56 &\cellgray 1.44(0.23)  \greendown
            & 2.97 &\cellgray 2.97(1.18)  \greendown
            & 1.48 &\cellgray 1.48(1.68)  \greendown
            & 1.55 &\cellgray 1.50(0.75)  \greendown
            & 0.46 &\cellgray 0.45(0.80)  \greendown
            \\
			& ResNet-110
            & 4.41 &\cellgray 0.96(0.23)  \greendown
            & 2.57 &\cellgray 2.34(1.15)  \greendown
            & 5.08 &\cellgray 2.16(0.22)  \greendown
            & 2.09 &\cellgray 2.09(1.19)  \greendown
            & 1.53 &\cellgray 1.54(1.26)  \redup
            & 1.88 &\cellgray 1.28(0.51)  \greendown
            & 0.98 &\cellgray 0.98(0.55)  \greendown
            \\
            & DenseNet-121
            & 4.51 &\cellgray 1.05(0.45)  \greendown
            & 1.52 &\cellgray 1.52(2.69)  \greendown
            & 5.10 &\cellgray 1.67(0.47)  \greendown
            & 1.87 &\cellgray 1.87(2.05)  \greendown
            & 1.31 &\cellgray 1.25(2.18)  \greendown
            & 1.23 &\cellgray 1.20(1.76)  \greendown
            & 0.57 &\cellgray 0.57(1.96)  \greendown
            \\
            & Wide-Resnet-26
            & 3.24 &\cellgray 1.35(0.28)  \greendown
            & 1.24 &\cellgray 1.24(2.08)  \greendown
            & 3.29 &\cellgray 1.30(0.28)  \greendown
            & 4.25 &\cellgray 4.25(1.74)  \greendown
            & 1.68 &\cellgray 1.68(2.02)  \greendown
            & 1.58 &\cellgray 1.58(2.20)  \greendown
            & 0.81 &\cellgray 0.81(2.12)  \greendown
            \\
			\midrule
			\multirow{4}{*}{CIFAR-100}
            & ResNet-50
            & 17.52 &\cellgray 3.98(0.60)  \greendown
            & 6.57 &\cellgray 3.96(2.11)  \greendown
            & 15.32 &\cellgray 4.79(0.72)  \greendown
            & 7.82 &\cellgray 7.82(3.67)  \greendown
            & 5.16 &\cellgray 4.79(2.43)  \greendown
            & 4.49 &\cellgray 3.83(2.17)  \greendown
            & 1.08 &\cellgray 1.01(2.23)  \greendown
            \\
			& ResNet-110
            & 19.06 &\cellgray 4.40(0.61)  \greendown
            & 7.87 &\cellgray 4.05(1.83)  \greendown
            & 19.14 &\cellgray 4.56(0.64)  \greendown
            & 11.04 &\cellgray 6.83(1.03)  \greendown
            & 8.66 &\cellgray 5.20(1.48)  \greendown
            & 8.55 &\cellgray 5.12(1.50)  \greendown
            & 2.90 &\cellgray 2.53(1.51)  \greendown
            \\
            & DenseNet-121
            & 20.99 &\cellgray 3.28(1.40)  \greendown
            & 5.22 &\cellgray 3.50(4.11)  \greendown
            & 19.11 &\cellgray 2.80(1.60)  \greendown
            & 12.87 &\cellgray 3.06(1.89)  \greendown
            & 4.14 &\cellgray 3.98(5.14)  \greendown
            & 3.70 &\cellgray 2.96(4.02)  \greendown
            & 1.81 &\cellgray 1.53(3.52)  \greendown
            \\
            & Wide-Resnet-26
            & 15.34 &\cellgray 4.38(0.98)  \greendown
            & 4.34 &\cellgray 3.11(2.24)  \greendown
            & 13.17 &\cellgray 4.18(1.07)  \greendown
            & 4.88 &\cellgray 4.88(3.10)  \greendown
            & 2.14 &\cellgray 1.70(3.09)  \greendown
            & 3.02 &\cellgray 1.90(2.47)  \greendown
            & 1.79 &\cellgray 1.18(2.30)  \greendown
            \\
			\bottomrule
		\end{tabular}%
	}
	\caption{\textbf{ECE$\downarrow$ before and after after feature clipping.}\quad  ECE is measured as a percentage, with lower values indicating better calibration. ECE is evaluated for different train-time calibration methods, both before (base) and after (+ours) feature clipping. The results are calculated with number of bins set as 15. The optimal $c$ is determined on the validation set, included in brackets.}
	\label{table:ece compare with training time methods}
\end{table*}

\subsubsection{Performance on Vision Transformer}
Although Vision Transformers do not end with a ReLU layer, the difference between HCE samples and LCE samples still exists, indicating that feature clipping can significantly influence HCE samples. As shown in Figure~\ref{fig:random_feature_distribution vit}, we take the mean of the absolute value of features for better visualization. The HCE samples show higher average feature values than LCE samples. However, the improvement is not as pronounced compared to CNN models. We show the ECE performance of ViT-L-16 in Table~\ref{table:ece on vit}. To investigate the reason, we count the number of overconfident samples, as shown in Table~\ref{table:overconfidence count for r50 and vit}. The number of samples with confidence larger than 0.8 is similar for both CNN and ViT. However, for samples with confidence greater than 0.95, CNN has significantly more samples than ViT. When examining samples with confidence greater than 0.99, CNN still has many samples, while ViT has none within this confidence range. This indicates that transformers face far fewer overconfidence issues compared to CNN models. Theoretically, clipping feature values results in a loss of information, increasing entropy and mitigating overconfidence problems. Since transformers exhibit fewer overconfidence problems compared to CNNs, our method has less impact on transformer-based models compared to CNNs. However, the difference in feature values among samples still exists, indicating significant potential for future improvements in transformer models.

\subsubsection{Compare with Train-time Calibration Methods}
We also compare the effectiveness of feature clipping when applied on top of various train-time calibration methods. Feature clipping consistently demonstrates improvement across all these train-time calibration methods and different models, as shown in Table~\ref{table:ece compare with training time methods}. On simpler datasets like CIFAR-10, models trained with ``maximum-entropy penalty" methods such as focal loss and label smoothing adequately address the overconfidence issue. These methods effectively mitigate the overconfidence problem, leaving little room for additional improvement through feature clipping. However, when applied to more complex datasets like CIFAR-100, these training losses may not entirely resolve the overconfidence problem, providing an opportunity for feature clipping to further alleviate this issue and enhance calibration. Feature clipping's ability to improve calibration in such scenarios underscores its potential as a valuable addition to existing training-time calibration methods.

\subsection{Ablation Study}
Feature clipping is a straightforward method that causes samples to lose information. We are interested in understanding how this loss of information affects various aspects of model performance. Therefore, we study its influence on accuracy, how hyperparameter $c$ affect performance, and its performance when applied to different layers.
\subsubsection{Does Feature Clipping Affect Accuracy?}
Although post-hoc methods do not change the model weights and can maintain prediction performance by keeping the original features, we are still interested in how the optimal clipping value affects accuracy. In Table~\ref{table:accuracy compare with other calibration methods}, we compare the prediction accuracy of different train-time calibration methods with our feature clipping method. The baseline column indicates the model trained with cross-entropy loss using the original features, while the FC column shows the results of applying our feature clipping on the baseline. All models are trained with the same training recipe, which is included in the Appendix. We observe that feature clipping does not significantly affect accuracy. Despite reducing the information contained in the feature representation, FC sometimes even improves accuracy. On the other hand, some train-time methods, such as Brier loss, can negatively impact accuracy in most cases. This suggests that while these methods aim to improve calibration, they may inadvertently reduce the model's ability to generalize, thereby lowering prediction accuracy. The detailed comparison of accuracy across different methods and datasets illustrates that our feature clipping method maintains competitive performance. 
\begin{table}[!ht]
	\centering
	\scriptsize
	\resizebox{\linewidth}{!}{%
    \begin{tabular}{ccccccccc}
        \toprule
        \textbf{Dataset} & \textbf{Model} & \textbf{Baseline} & \textbf{Brier Loss} & \textbf{LS-0.05} & \textbf{Focal Loss} & \textbf{FC (ours)}\\
                \midrule
        \multirow{3}{*}{CIFAR-10} 
        & ResNet-50         & 95.05 & 95.0  & 94.71 & 95.02 &94.93\\
        & ResNet-110        & 95.11 & 94.52  & 94.48 & 94.58 &95.01 \\
        & DenseNet-121      & 95.0 & 94.89  & 94.91 & 94.54 &95.13\\
        \midrule
        \multirow{3}{*}{CIFAR-100} 
        & ResNet-50         & 76.7 & 76.61  & 76.57 & 76.78 &76.74\\
        & ResNet-110        & 77.27 & 74.9  & 76.57 & 77.49  &77.06\\
        & DenseNet-121      & 75.48 & 76.25  & 75.95 & 77.33 &75.52\\

        \bottomrule
    \end{tabular}
	}
	\caption{\textbf{Accuracy$\uparrow$ for different train-time methods and feature clipping.}\quad Feature clipping does not impact prediction accuracy performance.}
	\label{table:accuracy compare with other calibration methods}
\end{table}

\begin{figure}[!ht]
    \centering
    \includegraphics[width=\linewidth]{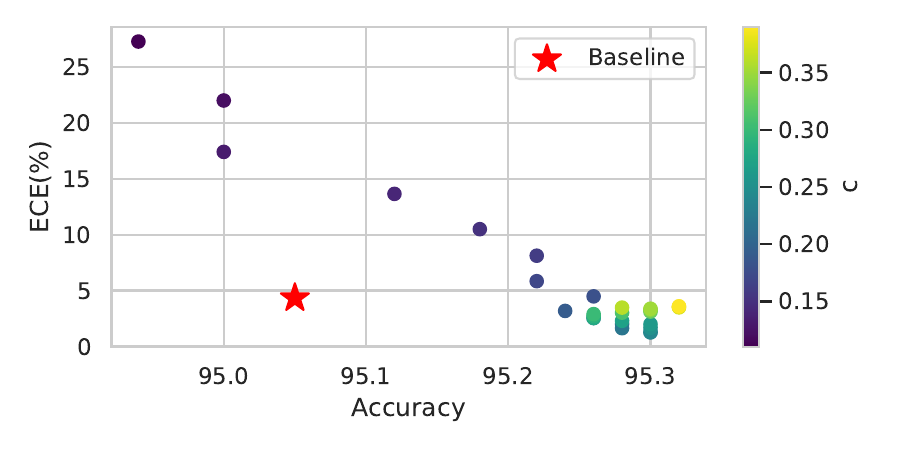}
    \caption{\textbf{Feature clipping at different value.}\quad Points to the right bottom cornor indicate better performance. The experiment is conducted on ResNet-50 on CIFAR-10.} 
    \label{fig: ablation study on clip value}
\end{figure}

\subsubsection{How does clip threshold affect performance?}
To test how the clipping threshold influences performance, we clip the features of a ResNet-50 trained with cross-entropy loss on CIFAR-10 using different clipping thresholds. We plot the resulting performance in terms of ECE and accuracy, as shown in Figure~\ref{fig: ablation study on clip value}. The red star indicates the performance of the original features. Generally, within a certain range (between 0.15 and 0.35 in this case), feature clipping does not significantly affect model accuracy. However, feature clipping can substantially influence calibration performance. For instance, a clipping value of 0.15 (point at the top left corner) achieves similar accuracy to the baseline but results in much worse calibration performance, with an ECE exceeding 25\%. With the optimal clipping value, the model can achieve an ECE as low as 1.10, as shown in Table~\ref{table:ece compare with posthoc methods}. We believe the reason feature clipping has a larger influence on calibration is that excessive clipping may significantly increase entropy, affecting correct predictions with high confidence. As a result, the model faces underconfidence, leading to a large ECE.

\subsubsection{How about clipping on other layer?}
To test the influence of applying feature clipping on different layers, we applied feature clipping to various layers of ResNet-50 and evaluated the performance, as shown in Table~\ref{table:clip on other layer}. When clipping the last layer (logits before SoftMax) and the penultimate layer (features), feature clipping achieves excellent calibration performance without significantly affecting accuracy. However, applying clipping to earlier layers considerably degrades both calibration and prediction performance. We hypothesize that this is because clipping at earlier stages has a larger impact on information flow, thereby significantly harming model performance. In contrast, clipping on logits shows competitive results and merits further investigation in future studies.
\begin{table}
\centering
\begin{tabular}[width=\linewidth]{l|c|c}
\toprule
& \textbf{ECE (c)} & \textbf{Accuracy} \\ \midrule
No Clipping (baseline) & 4.34 & 95.05 \\
Last Layer (logits) & 1.21 (4.98) & 94.62 \\
Penultimate Layer (ours) & 1.10 (0.23) & 94.93 \\
Antepenultimate Layer & 4.42 (0.35) & 94.95 \\ 
Pre-antepenultimate Layer & 14.42 (0.41) & 78.12 \\ 
\bottomrule
\end{tabular}
\caption{\textbf{ECE and accuracy when clipping outputs at different layers.} Clipping features and logits provides competitive calibration performance. The experiment is conducted on ResNet-50 on CIFAR-10.}
\label{table:clip on other layer}
\end{table}

\section{Conclusion}
In conclusion, our proposed feature clipping method demonstrates substantial improvements in model calibration across various datasets and models. FC effectively reduces overconfidence in predictions, enhancing calibration performance while maintaining accuracy. Despite its simplicity, FC achieves state-of-the-art calibration performance and provides a solid foundation for future research on feature-based calibration. However, there are several limitations and opportunities for improvement. The performance on transformer models, for instance, can be further improved. Future work should focus on developing more sophisticated methods, such as starting with a better threshold and conducting faster hyperparameter tuning. Additionally, exploring ways to find optimal clipping values using feature statistics and employing smoothed or adaptive thresholds instead of fixed ones are promising directions. These enhancements will potentially lead to even better calibration. Furthermore, investigating the impact of feature clipping on different neural network architectures and understanding its effects on various types of data can provide deeper insights. Our method serves as a strong baseline for feature-based calibration, and we believe that future developments can build upon this foundation to achieve even greater calibration improvements.


\clearpage
\clearpage
\bibliography{aaai25}

\begin{thebibliography}{38}
\providecommand{\natexlab}[1]{#1}

\bibitem[{Brier(1950)}]{brier1950verification}
Brier, G.~W. 1950.
\newblock Verification of forecasts expressed in terms of probability.
\newblock \emph{Monthly weather review}, 78(1): 1--3.

\bibitem[{Burkardt(2014)}]{burkardt2014truncated}
Burkardt, J. 2014.
\newblock The truncated normal distribution.
\newblock \emph{Department of Scientific Computing Website, Florida State University}, 1(35): 58.

\bibitem[{Chen et~al.(2018)Chen, Sahiner, Samuelson, Pezeshk, and Petrick}]{chen2018calibration}
Chen, W.; Sahiner, B.; Samuelson, F.; Pezeshk, A.; and Petrick, N. 2018.
\newblock Calibration of medical diagnostic classifier scores to the probability of disease.
\newblock \emph{Statistical methods in medical research}, 27(5): 1394--1409.

\bibitem[{Deng et~al.(2009)Deng, Dong, Socher, Li, Li, and Fei-Fei}]{deng2009imagenet}
Deng, J.; Dong, W.; Socher, R.; Li, L.-J.; Li, K.; and Fei-Fei, L. 2009.
\newblock Imagenet: A large-scale hierarchical image database.
\newblock In \emph{2009 IEEE conference on computer vision and pattern recognition}, 248--255. Ieee.

\bibitem[{Dosovitskiy et~al.(2020)Dosovitskiy, Beyer, Kolesnikov, Weissenborn, Zhai, Unterthiner, Dehghani, Minderer, Heigold, Gelly et~al.}]{dosovitskiy2020image}
Dosovitskiy, A.; Beyer, L.; Kolesnikov, A.; Weissenborn, D.; Zhai, X.; Unterthiner, T.; Dehghani, M.; Minderer, M.; Heigold, G.; Gelly, S.; et~al. 2020.
\newblock An image is worth 16x16 words: Transformers for image recognition at scale.
\newblock \emph{arXiv preprint arXiv:2010.11929}.

\bibitem[{Feng et~al.(2019)Feng, Rosenbaum, Glaeser, Timm, and Dietmayer}]{feng2019can}
Feng, D.; Rosenbaum, L.; Glaeser, C.; Timm, F.; and Dietmayer, K. 2019.
\newblock Can we trust you? on calibration of a probabilistic object detector for autonomous driving.
\newblock \emph{arXiv preprint arXiv:1909.12358}.

\bibitem[{Frenkel et~al.(2021)Frenkel, Goldberger, Goldberger, and Goldberger}]{frenkel2021network}
Frenkel, L.; Goldberger, J.; Goldberger, J.; and Goldberger, J. 2021.
\newblock Network calibration by class-based temperature scaling.
\newblock In \emph{2021 29th European Signal Processing Conference (EUSIPCO)}, 1486--1490. IEEE.

\bibitem[{Ghosh, Schaaf, and Gormley(2022)}]{ghosh2022adafocal}
Ghosh, A.; Schaaf, T.; and Gormley, M. 2022.
\newblock Adafocal: Calibration-aware adaptive focal loss.
\newblock \emph{Advances in Neural Information Processing Systems}, 35: 1583--1595.

\bibitem[{Guo et~al.(2017)Guo, Pleiss, Sun, and Weinberger}]{guo2017calibration}
Guo, C.; Pleiss, G.; Sun, Y.; and Weinberger, K.~Q. 2017.
\newblock On calibration of modern neural networks.
\newblock In \emph{International conference on machine learning}, 1321--1330. PMLR.

\bibitem[{He et~al.(2016)He, Zhang, Ren, and Sun}]{he2016deep}
He, K.; Zhang, X.; Ren, S.; and Sun, J. 2016.
\newblock Deep residual learning for image recognition.
\newblock In \emph{Proceedings of the IEEE conference on computer vision and pattern recognition}, 770--778.

\bibitem[{H{\'e}bert-Johnson et~al.(2018)H{\'e}bert-Johnson, Kim, Reingold, and Rothblum}]{hebert2018multicalibration}
H{\'e}bert-Johnson, U.; Kim, M.; Reingold, O.; and Rothblum, G. 2018.
\newblock Multicalibration: Calibration for the (computationally-identifiable) masses.
\newblock In \emph{International Conference on Machine Learning}, 1939--1948. PMLR.

\bibitem[{Hendrycks et~al.(2019)Hendrycks, Mu, Cubuk, Zoph, Gilmer, and Lakshminarayanan}]{hendrycks2019augmix}
Hendrycks, D.; Mu, N.; Cubuk, E.~D.; Zoph, B.; Gilmer, J.; and Lakshminarayanan, B. 2019.
\newblock Augmix: A simple data processing method to improve robustness and uncertainty.
\newblock \emph{arXiv preprint arXiv:1912.02781}.

\bibitem[{Howard et~al.(2017)Howard, Zhu, Chen, Kalenichenko, Wang, Weyand, Andreetto, and Adam}]{howard2017mobilenets}
Howard, A.~G.; Zhu, M.; Chen, B.; Kalenichenko, D.; Wang, W.; Weyand, T.; Andreetto, M.; and Adam, H. 2017.
\newblock Mobilenets: Efficient convolutional neural networks for mobile vision applications.
\newblock \emph{arXiv preprint arXiv:1704.04861}.

\bibitem[{Huang et~al.(2017)Huang, Liu, Van Der~Maaten, and Weinberger}]{huang2017densely}
Huang, G.; Liu, Z.; Van Der~Maaten, L.; and Weinberger, K.~Q. 2017.
\newblock Densely connected convolutional networks.
\newblock In \emph{Proceedings of the IEEE conference on computer vision and pattern recognition}, 4700--4708.

\bibitem[{Ilyas et~al.(2019)Ilyas, Santurkar, Tsipras, Engstrom, Tran, and Madry}]{ilyas2019adversarial}
Ilyas, A.; Santurkar, S.; Tsipras, D.; Engstrom, L.; Tran, B.; and Madry, A. 2019.
\newblock Adversarial examples are not bugs, they are features.
\newblock \emph{Advances in neural information processing systems}, 32.

\bibitem[{Krishnan and Tickoo(2020)}]{krishnan2020improving}
Krishnan, R.; and Tickoo, O. 2020.
\newblock Improving model calibration with accuracy versus uncertainty optimization.
\newblock \emph{Advances in Neural Information Processing Systems}, 33: 18237--18248.

\bibitem[{Krizhevsky, Hinton et~al.(2009)}]{krizhevsky2009learning}
Krizhevsky, A.; Hinton, G.; et~al. 2009.
\newblock Learning multiple layers of features from tiny images.
\newblock \emph{arXiv preprint arXiv:2010.11929}.

\bibitem[{Kull et~al.(2019)Kull, Perello~Nieto, K{\"a}ngsepp, Silva~Filho, Song, and Flach}]{kull2019beyond}
Kull, M.; Perello~Nieto, M.; K{\"a}ngsepp, M.; Silva~Filho, T.; Song, H.; and Flach, P. 2019.
\newblock Beyond temperature scaling: Obtaining well-calibrated multi-class probabilities with dirichlet calibration.
\newblock \emph{Advances in neural information processing systems}, 32.

\bibitem[{Kumar et~al.(2018)Kumar, Sarawagi, Jain, and Jain}]{kumar2018trainable}
Kumar, A.; Sarawagi, S.; Jain, U.; and Jain, U. 2018.
\newblock Trainable calibration measures for neural networks from kernel mean embeddings.
\newblock In \emph{International Conference on Machine Learning}, 2805--2814. PMLR.

\bibitem[{Minderer et~al.(2021)Minderer, Djolonga, Romijnders, Hubis, Zhai, Houlsby, Tran, and Lucic}]{minderer2021revisiting}
Minderer, M.; Djolonga, J.; Romijnders, R.; Hubis, F.; Zhai, X.; Houlsby, N.; Tran, D.; and Lucic, M. 2021.
\newblock Revisiting the calibration of modern neural networks.
\newblock \emph{Advances in Neural Information Processing Systems}, 34: 15682--15694.

\bibitem[{Mukhoti et~al.(2020)Mukhoti, Kulharia, Sanyal, Golodetz, Torr, and Dokania}]{mukhoti2020calibrating}
Mukhoti, J.; Kulharia, V.; Sanyal, A.; Golodetz, S.; Torr, P.; and Dokania, P. 2020.
\newblock Calibrating deep neural networks using focal loss.
\newblock \emph{Advances in Neural Information Processing Systems}, 33: 15288--15299.

\bibitem[{M{\"u}ller et~al.(2019)M{\"u}ller, Kornblith, Hinton, and Hinton}]{muller2019does}
M{\"u}ller, R.; Kornblith, S.; Hinton, G.~E.; and Hinton, G.~E. 2019.
\newblock When does label smoothing help?
\newblock \emph{Advances in neural information processing systems}, 32.

\bibitem[{Pereyra et~al.(2017)Pereyra, Tucker, Chorowski, Kaiser, and Hinton}]{pereyra2017regularizing}
Pereyra, G.; Tucker, G.; Chorowski, J.; Kaiser, {\L}.; and Hinton, G. 2017.
\newblock Regularizing neural networks by penalizing confident output distributions.
\newblock \emph{arXiv preprint arXiv:1701.06548}.

\bibitem[{Politis(1991)}]{politis1991entropy}
Politis, D.~N. 1991.
\newblock On the Entropy of a Mixture Distribution.
\newblock Technical Report 91-67, Purdue University.

\bibitem[{Socci, Lee, and Seung(1997)}]{socci1997rectified}
Socci, N.; Lee, D.; and Seung, H.~S. 1997.
\newblock The rectified Gaussian distribution.
\newblock \emph{Advances in neural information processing systems}, 10.

\bibitem[{Tao et~al.(2023{\natexlab{a}})Tao, Dong, Liu, Sun, and Xu}]{tao2023calibrating}
Tao, L.; Dong, M.; Liu, D.; Sun, C.; and Xu, C. 2023{\natexlab{a}}.
\newblock Calibrating a deep neural network with its predecessors.
\newblock In \emph{Proceedings of the Thirty-Second International Joint Conference on Artificial Intelligence}, 4271--4279.

\bibitem[{Tao et~al.(2023{\natexlab{b}})Tao, Dong, Xu, and Xu}]{tao2023dual}
Tao, L.; Dong, M.; Xu, C.; and Xu, C. 2023{\natexlab{b}}.
\newblock Dual focal loss for calibration.
\newblock In \emph{International Conference on Machine Learning}, 33833--33849. PMLR.

\bibitem[{Tao et~al.(2023{\natexlab{c}})Tao, Zhu, Guo, Dong, and Xu}]{tao2023benchmark}
Tao, L.; Zhu, Y.; Guo, H.; Dong, M.; and Xu, C. 2023{\natexlab{c}}.
\newblock A benchmark study on calibration.
\newblock \emph{arXiv preprint arXiv:2308.11838}.

\bibitem[{Tomani et~al.(2022)Tomani, Cremers, Buettner, and Sun}]{tomani2022parameterized}
Tomani, C.; Cremers, D.; Buettner, F.; and Sun, Y. 2022.
\newblock Parameterized temperature scaling for boosting the expressive power in post-hoc uncertainty calibration.
\newblock In \emph{European Conference on Computer Vision}, 555--569. Springer.

\bibitem[{Wang, Feng, and Zhang(2021)}]{wang2021rethinking}
Wang, D.-B.; Feng, L.; and Zhang, M.-L. 2021.
\newblock Rethinking calibration of deep neural networks: Do not be afraid of overconfidence.
\newblock \emph{Advances in Neural Information Processing Systems}, 34: 11809--11820.

\bibitem[{Wang et~al.(2023)Wang, Li, Zhao, Heng, and Zhang}]{wang2023pitfall}
Wang, D.-B.; Li, L.; Zhao, P.; Heng, P.-A.; and Zhang, M.-L. 2023.
\newblock On the pitfall of mixup for uncertainty calibration.
\newblock In \emph{Proceedings of the IEEE/CVF Conference on Computer Vision and Pattern Recognition}, 7609--7618.

\bibitem[{Xiong et~al.(2023)Xiong, Deng, Koh, Wu, Li, Xu, and Hooi}]{xiong2023proximity}
Xiong, M.; Deng, A.; Koh, P. W.~W.; Wu, J.; Li, S.; Xu, J.; and Hooi, B. 2023.
\newblock Proximity-informed calibration for deep neural networks.
\newblock \emph{Advances in Neural Information Processing Systems}, 36: 68511--68538.

\bibitem[{Yang et~al.(2024)Yang, Zhan, Gan, and Sun}]{yang2024beyond}
Yang, J.-Q.; Zhan, D.-C.; Gan, L.; and Sun, Y. 2024.
\newblock Beyond probability partitions: Calibrating neural networks with semantic aware grouping.
\newblock \emph{Advances in Neural Information Processing Systems}, 36.

\bibitem[{Zadrozny and Elkan(2001)}]{zadrozny2001obtaining}
Zadrozny, B.; and Elkan, C. 2001.
\newblock Obtaining calibrated probability estimates from decision trees and naive bayesian classifiers.
\newblock In \emph{Icml}, volume~1, 609--616.

\bibitem[{Zadrozny and Elkan(2002)}]{zadrozny2002transforming}
Zadrozny, B.; and Elkan, C. 2002.
\newblock Transforming classifier scores into accurate multiclass probability estimates.
\newblock In \emph{Proceedings of the eighth ACM SIGKDD international conference on Knowledge discovery and data mining}, 694--699.

\bibitem[{Zagoruyko and Komodakis(2016)}]{zagoruyko2016wide}
Zagoruyko, S.; and Komodakis, N. 2016.
\newblock Wide residual networks.
\newblock \emph{arXiv preprint arXiv:1605.07146}.

\bibitem[{Zhang et~al.(2020)Zhang, Kailkhura, Han, and Sun}]{zhang2020mix}
Zhang, J.; Kailkhura, B.; Han, T. Y.-J.; and Sun, Y. 2020.
\newblock Mix-n-match: Ensemble and compositional methods for uncertainty calibration in deep learning.
\newblock In \emph{International conference on machine learning}, 11117--11128. PMLR.

\bibitem[{Zhang et~al.(2022)Zhang, Deng, Kawaguchi, and Zou}]{zhang2022and}
Zhang, L.; Deng, Z.; Kawaguchi, K.; and Zou, J. 2022.
\newblock When and how mixup improves calibration.
\newblock In \emph{International Conference on Machine Learning}, 26135--26160. PMLR.

\end{thebibliography}
\clearpage

\section{Additional Resources}
\section{Detail Derivation \\of Entropy Difference Monotonicity}
\subsubsection{Entropy Difference}Then, the Shannon entropy difference between features before and after clipping is given by

\begin{align}
\Delta H =&-\left(1.5-\Phi\left(\frac{c}{\sigma}\right)\right)\log\left(1.5-\Phi\left(\frac{c}{\sigma}\right)\right)\notag\\
    &-\left(\Phi\left(\frac{c}{\sigma}\right)-0.5\right)\log\left(\Phi\left(\frac{c}{\sigma}\right)-0.5\right)\notag\\
    &-0.5\log\left(\frac{0.5}{1.5-\Phi\left(\frac{c}{\sigma}\right)}\right)\notag\\
    &-\left(1-\Phi\left(\frac{c}{\sigma}\right)\right)\log\left(\frac{1-\Phi\left(\frac{c}{\sigma}\right)}{1.5-\Phi\left(\frac{c}{\sigma}\right)}\right)\notag\\
    & + (\Phi\left(\frac{c}{\sigma}\right)-0.5)H_c(\tilde{x}; 0, \sigma, 0, c) \notag\\
    &+\log(\frac{1}{2})+\frac{1}{2}\log(\sqrt{\pi e \sigma})\notag\\
    =&-\left(1.5-\Phi\left(\frac{c}{\sigma}\right)\right)\log\left(1.5-\Phi\left(\frac{c}{\sigma}\right)\right)\notag\\
    &-\left(\Phi\left(\frac{c}{\sigma}\right)-0.5\right)\log\left(\Phi\left(\frac{c}{\sigma}\right)-0.5\right)\notag\\
&-0.5\log\left(0.5\right)+0.5\log\left(1.5-\Phi\left(\frac{c}{\sigma}\right)\right)\notag\\
    &-\left(1-\Phi\left(\frac{c}{\sigma}\right)\right)\log\left(\frac{1-\Phi\left(\frac{c}{\sigma}\right)}{1.5-\Phi\left(\frac{c}{\sigma}\right)}\right)\notag\\
    & + (\Phi\left(\frac{c}{\sigma}\right)-0.5)H_c(\tilde{x}; 0, \sigma, 0, c)\notag\\
    &+\log(\frac{1}{2})+\frac{1}{2}\log(\sqrt{\pi e \sigma})\notag\\
\label{eq:entropy difference of clipped feature}
\end{align}
Simplify:
\begin{align}
    \Delta H =&-\left(\Phi\left(\frac{c}{\sigma}\right)-0.5\right)\log\left(\Phi\left(\frac{c}{\sigma}\right)-0.5\right)\notag\\
&+0.5\log\left(0.5\right)\notag\\
    &-\left(1-\Phi\left(\frac{c}{\sigma}\right)\right)\log\left(1-\Phi\left(\frac{c}{\sigma}\right)\right)\notag\\
    & + (\Phi\left(\frac{c}{\sigma}\right)-0.5)H_c(\tilde{x}; 0, \sigma, 0, c)\notag\\
    &+\frac{1}{2}\log(\sqrt{\pi e \sigma})
\end{align}

\subsubsection{Determine the Monotonicity}To calculate the monotonicity of $\Delta H$ with respect to $\sigma$. We need to calculate $\frac{d\Delta H}{d\sigma}$ with respect to.

For the first term:

\[
- \left( \Phi \left( \frac{c}{\sigma} \right) - 0.5 \right) \log \left( \Phi \left( \frac{c}{\sigma} \right) - 0.5 \right)
\]

Let $ u = \Phi \left( \frac{c}{\sigma} \right) - 0.5 $:

\[
\frac{d}{d\sigma} \left[ - u \log u \right] = - \left( \log u \frac{du}{d\sigma} + \frac{u}{u} \frac{du}{d\sigma} \right)
\]

\[
\frac{du}{d\sigma} = \phi \left( \frac{c}{\sigma} \right) \left( -\frac{c}{\sigma^2} \right) = -\phi \left( \frac{c}{\sigma} \right) \frac{c}{\sigma^2}
\]

Thus,

\[
\frac{d}{d\sigma} \left[ - \left( \Phi \left( \frac{c}{\sigma} \right) - 0.5 \right) \log \left( \Phi \left( \frac{c}{\sigma} \right) - 0.5 \right) \right]
\]

\[
= \left( \log \left( \Phi \left( \frac{c}{\sigma} \right) - 0.5 \right) + 1 \right) \phi \left( \frac{c}{\sigma} \right) \frac{c}{\sigma^2}
\]

The second term is constant, so its derivative is 0.

For the third term:

\[
- \left( 1 - \Phi \left( \frac{c}{\sigma} \right) \right) \log \left( 1 - \Phi \left( \frac{c}{\sigma} \right) \right)
\]

Let $ v = 1 - \Phi \left( \frac{c}{\sigma} \right) $:

\[
\frac{d}{d\sigma} \left[ - v \log v \right] = - \left( \log v \frac{dv}{d\sigma} + \frac{v}{v} \frac{dv}{d\sigma} \right)
\]

\[
\frac{dv}{d\sigma} = -\phi \left( \frac{c}{\sigma} \right) \left( -\frac{c}{\sigma^2} \right) = \phi \left( \frac{c}{\sigma} \right) \frac{c}{\sigma^2}
\]

Thus,

\[
\frac{d}{d\sigma} \left[ - \left( 1 - \Phi \left( \frac{c}{\sigma} \right) \right) \log \left( 1 - \Phi \left( \frac{c}{\sigma} \right) \right) \right] 
\]

\[
= - \left( \log \left( 1 - \Phi \left( \frac{c}{\sigma} \right) \right) + 1 \right) \phi \left( \frac{c}{\sigma} \right) \frac{c}{\sigma^2}
\]

For the fourth term:

\[
\left( \Phi \left( \frac{c}{\sigma} \right) - 0.5 \right) H_c(\tilde{x}; 0, \sigma, 0, c)
\]

Let $ u = \Phi \left( \frac{c}{\sigma} \right) - 0.5 $:

\[
\frac{d}{d\sigma} \left[ u H_c(\tilde{x}; 0, \sigma, 0, c) \right]
\]

\[
= \frac{d}{d\sigma} \left[ u \right] H_c(\tilde{x}; 0, \sigma, 0, c) + u \frac{d}{d\sigma} \left[ H_c(\tilde{x}; 0, \sigma, 0, c) \right]
\]

\[
\frac{d}{d\sigma} \left[ u \right] = \phi \left( \frac{c}{\sigma} \right) \left( -\frac{c}{\sigma^2} \right) = -\phi \left( \frac{c}{\sigma} \right) \frac{c}{\sigma^2}
\]

Thus,

\[
\frac{d}{d\sigma} \left[ u H_c(\tilde{x}; 0, \sigma, 0, c) \right]
\]

\[
= - \phi \left( \frac{c}{\sigma} \right) \frac{c}{\sigma^2} H_c(\tilde{x}; 0, \sigma, 0, c) + u \frac{d}{d\sigma} \left[ H_c(\tilde{x}; 0, \sigma, 0, c) \right]
\]

Where:

\[
\frac{d}{d\sigma} \left[ H_c(\tilde{x}; 0, \sigma, 0, c) \right]
\]

This term needs to be evaluated based on the definition of $H_c$. We can use the given expression for $H_c$:

\[
H_c(x; \mu, \sigma, a, b) = \log(\sqrt{2\pi e \sigma^2 Z}) + \frac{\alpha \phi(\alpha) - \beta \phi(\beta)}{2Z}
\]

For simplicity, let $\alpha = \frac{0}{\sigma} = 0$ and $\beta = \frac{c}{\sigma}$ and $Z = \Phi(\frac{c}{\sigma}) - \Phi(0) = \Phi(\frac{c}{\sigma}) - 0 = \Phi(\frac{c}{\sigma})$.

\[
\frac{d}{d\sigma} \left[ \log(\sqrt{2 \pi e \sigma^2 \Phi(\frac{c}{\sigma})}) + \frac{\left( -\beta \phi(\beta) \right)}{2 \Phi(\beta)} \right]
\]

Differentiating the first term:
\[
\frac{d}{d\sigma} \log(\sqrt{2 \pi e \sigma^2 \Phi(\frac{c}{\sigma})})
\]
\[
= \frac{d}{d\sigma} \log(\sigma) +\frac{d}{d\sigma}\frac{1}{2} \log(\Phi(\frac{c}{\sigma}))= \frac{1}{\sigma}-\frac{ c\phi(\frac{c}{\sigma})}{2\sigma^2 \Phi(\frac{c}{\sigma})} 
\]

For the second term in $H_c$:

\[
\frac{d}{d\sigma} \left[ \frac{- \beta \phi(\beta)}{2 \Phi(\beta)} \right] = \frac{d}{d\sigma} \left[ \frac{- \frac{c}{\sigma} \phi(\frac{c}{\sigma})}{2 \Phi(\frac{c}{\sigma})} \right]
\]

Applying the quotient rule:

\[
= \frac{\left( - \frac{d}{d\sigma} \left[ \frac{c}{\sigma} \phi(\frac{c}{\sigma}) \right] \right) \cdot 2 \Phi(\frac{c}{\sigma}) - \left( - \frac{c}{\sigma} \phi(\frac{c}{\sigma}) \right) \cdot \frac{d}{d\sigma} \left[ 2 \Phi(\frac{c}{\sigma}) \right]}{(2 \Phi(\frac{c}{\sigma}))^2}
\]

\[
= \frac{\left(-\phi\left(\frac{c}{\sigma}\right) \frac{c (c^2 - \sigma^2)}{\sigma^4}
 \right) \cdot 2 \Phi(\frac{c}{\sigma}) - \frac{c}{\sigma} \phi(\frac{c}{\sigma}) \cdot 2 \phi(\frac{c}{\sigma}) \cdot \frac{c}{\sigma^2}}{4 (\Phi(\frac{c}{\sigma}))^2}
\]

\[
=\frac{-\phi\left(\frac{c}{\sigma}\right)}{2 (\Phi(\frac{c}{\sigma}))^2} \left( \frac{c (c^2 - \sigma^2)}{\sigma^4} \Phi\left(\frac{c}{\sigma}\right) + \frac{c^2}{\sigma^3} \phi\left(\frac{c}{\sigma}\right) \right)
\]

Hence, the derivative of $H_c$ with respect to $\sigma$ is:

\[
\frac{1}{\sigma}-\frac{ c\phi(\frac{c}{\sigma})}{2\sigma^2 \Phi(\frac{c}{\sigma})} + \frac{-\phi\left(\frac{c}{\sigma}\right)}{2 (\Phi(\frac{c}{\sigma}))^2} \left( \frac{c (c^2 - \sigma^2)}{\sigma^4} \Phi\left(\frac{c}{\sigma}\right) + \frac{c^2}{\sigma^3} \phi\left(\frac{c}{\sigma}\right) \right)
\]

For the fifth term:

\[
\frac{1}{2} \log (\sqrt{\pi e \sigma})
\]

\[
\frac{d}{d\sigma} \left[ \frac{1}{2} \log (\sqrt{\pi e \sigma}) \right] = \frac{1}{2} \frac{1}{\sqrt{\pi e \sigma}} \cdot \frac{\pi e}{2\sqrt{\pi e \sigma}} =\frac{1}{4\sigma}
\]

Combining all these terms, we have the final derivative of $\Delta H$ with respect to $\sigma$:

\begin{align}
&\frac{d (\Delta H)}{d\sigma}=\left( \log \left( \Phi \left( \frac{c}{\sigma} \right) - 0.5 \right) + 1 \right) \phi \left( \frac{c}{\sigma} \right) \frac{c}{\sigma^2}  \notag\\
&- \left( \log \left( 1 - \Phi \left( \frac{c}{\sigma} \right) \right) + 1 \right) \phi \left( \frac{c}{\sigma} \right) \frac{c}{\sigma^2}  \notag\\
&+\frac{1}{\sigma}-\frac{ c\phi(\frac{c}{\sigma})}{2\sigma^2 \Phi(\frac{c}{\sigma})}    \notag\\
&+\frac{-\phi\left(\frac{c}{\sigma}\right)}{2 (\Phi(\frac{c}{\sigma}))^2} \left( \frac{c (c^2 - \sigma^2)}{\sigma^4} \Phi\left(\frac{c}{\sigma}\right) + \frac{c^2}{\sigma^3} \phi\left(\frac{c}{\sigma}\right) \right)  \notag\\
&+\frac{1}{4\sigma}  \notag\\
&=\log \left( \frac{\Phi \left( \frac{c}{\sigma} \right) - 0.5}{ 1 - \Phi \left( \frac{c}{\sigma} \right)} \right)\cdot\phi \left( \frac{c}{\sigma} \right) \frac{c}{\sigma^2}  \notag\\
&+\frac{5}{4\sigma}-\frac{ c\phi(\frac{c}{\sigma})}{2\sigma^2 \Phi(\frac{c}{\sigma})}    \notag\\
&+\frac{-\phi\left(\frac{c}{\sigma}\right)}{2 (\Phi(\frac{c}{\sigma}))^2} \left( \frac{c (c^2 - \sigma^2)}{\sigma^4} \Phi\left(\frac{c}{\sigma}\right) + \frac{c^2}{\sigma^3} \phi\left(\frac{c}{\sigma}\right) \right)  \notag\\
\end{align}

\begin{figure}[!ht]
    \centering
    \includegraphics[width=\linewidth]{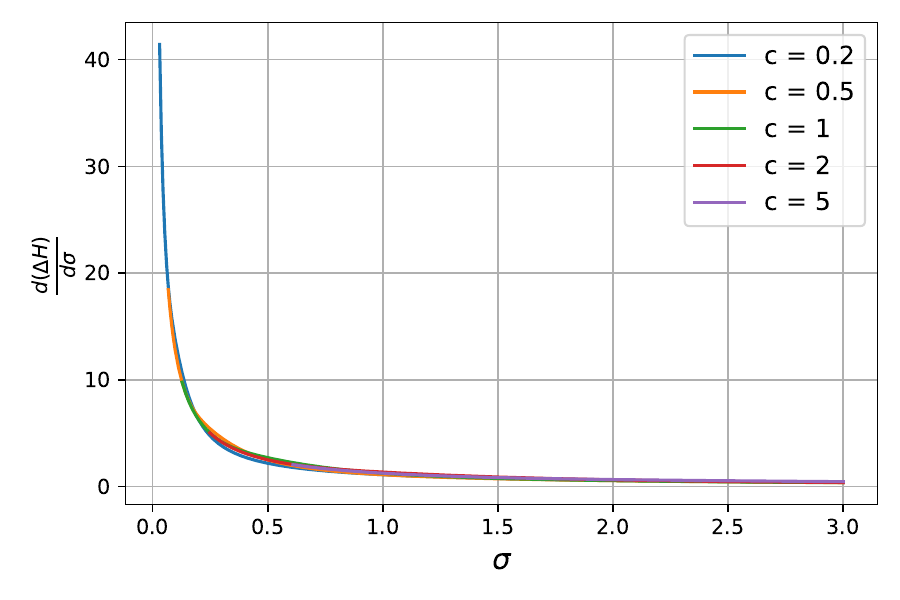}
    \caption{\textbf{The derivative of $\Delta H$}~Different color indicates different choice of clipping hyperparameter $c$.}
    \label{fig:derivative of dalta H}
\end{figure}

However, it is non-trivial to calculate the analytical solution of $\frac{d(\Delta H)}{d\sigma} = 0$ to determine the monotonicity of $\Delta H$ with respect to $\sigma$. Instead, we visualize $\frac{d(\Delta H)}{d\sigma}$ with respect to $\sigma$ at different $c$ to numerically study the monotonicity. As shown in Figure~\ref{fig:derivative of dalta H}, $\frac{d(\Delta H)}{d\sigma}$ is positive for all $\sigma>0$. In other words, $\Delta H$ is monotonically increase with respect to $\sigma$ for all $\sigma>0$. We observe that, in practice, $\sigma_{LCE}<\sigma_{HCE}$, as shown in Figure~\ref{fig:feature value histogram density}. Thus, we can derive the conclusion that \textit{HCE samples suffer larger entropy difference compared to LCE samples after feature clipping, i.e. $\Delta H_{HCE}>\Delta H_{LCE}$}.

\begin{figure*}
    \centering
    \includegraphics[width=\linewidth]{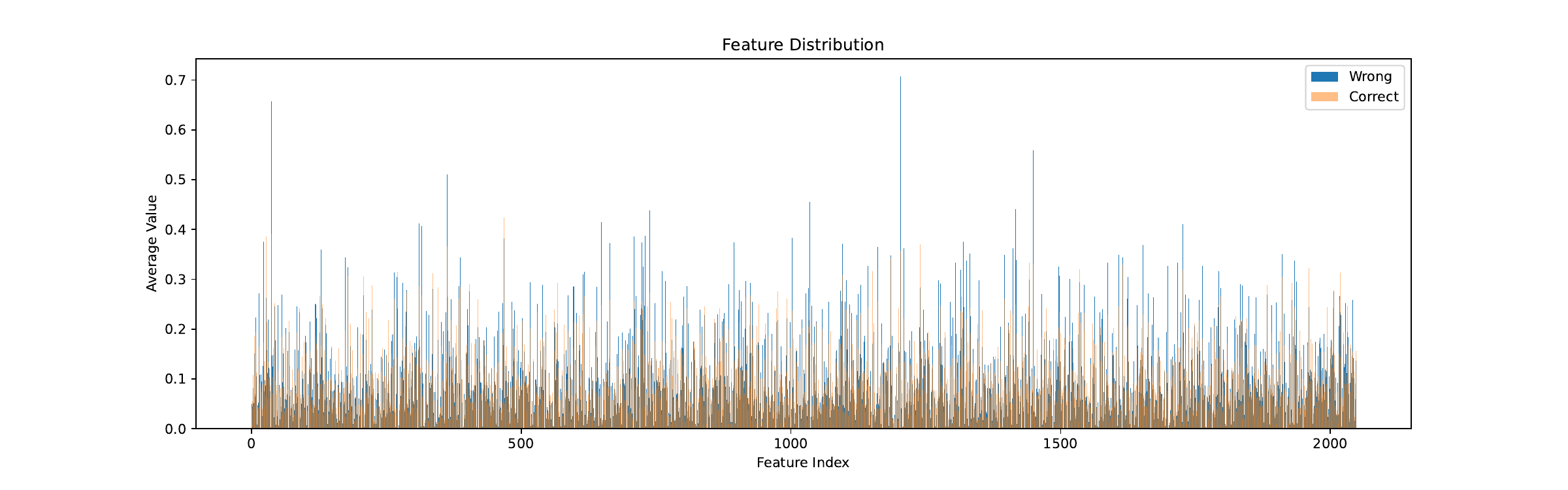}
    \caption{\textbf{Average feature value of samples with high or low calibration error for feature values in all 2048 dimension.}~The experiment is conducted on ResNet-50 on CIFAR-10.}
    \label{fig:full_feature_distribution}
\end{figure*}

\begin{table*}
	\centering
	\scriptsize
	\resizebox{\linewidth}{!}{%
		\begin{tabular}{ccccccccccccccc}
			\toprule
			\textbf{Dataset} & \textbf{Model} & \multicolumn{2}{c}{\textbf{Original Feature}} &\multicolumn{2}{c}{\textbf{TS}} &\multicolumn{2}{c}{\textbf{ETS}} &
			\multicolumn{2}{c}{\textbf{PTS}} &
   \multicolumn{2}{c}{\textbf{CTS}} &
   \multicolumn{2}{c}{\textbf{Group Calibration}} & 
            \\
			\textbf{} & \textbf{} & \multicolumn{2}{c}{ } &
			\multicolumn{2}{c}{\citep{guo2017calibration}} &
   \multicolumn{2}{c}{\citep{zhang2020mix}} &
   \multicolumn{2}{c}{\citep{tomani2022parameterized}} &
    \multicolumn{2}{c}{\citep{frenkel2021network}} &
    \multicolumn{2}{c}{\citep{yang2024beyond}} &

           \\
			&& base & +ours(c) & base & +ours & base & +ours & base & +ours & base & 
			+ours & base & +ours \\
			\midrule
			\multirow{3}{*}{CIFAR-10} 
            & ResNet-50&4.33&\cellgray1.74(0.23) \greendown&2.14&\cellgray1.77 \greendown&2.14&\cellgray1.77 \greendown&2.14&\cellgray1.75 \greendown&1.73&\cellgray1.75 \redup&1.23&\cellgray0.35 \greendown\\
			& ResNet-110&4.40&\cellgray1.65(0.23) \greendown&1.91&\cellgray1.66 \greendown&1.91&\cellgray1.66 \greendown&1.92&\cellgray1.66 \greendown&1.31&\cellgray1.66 \redup&0.87&\cellgray1.94 \redup\\
            & DenseNet-121&4.49&\cellgray1.52(0.45) \greendown&2.12&\cellgray1.53 \greendown&2.12&\cellgray1.53 \greendown&2.12&\cellgray1.53 \greendown&1.73&\cellgray1.53 \greendown&1.28&\cellgray2.48 \redup\\
			\midrule
			\multirow{3}{*}{CIFAR-100}
            & ResNet-50&17.52&\cellgray4.01(0.6) \greendown&5.72&\cellgray4.26 \greendown&5.70&\cellgray4.29 \greendown&5.66&\cellgray4.37 \greendown&5.77&\cellgray4.37 \greendown&3.49&\cellgray1.78 \greendown\\
			& ResNet-110&19.06&\cellgray4.80(0.61) \greendown&6.28&\cellgray5.22 \greendown&6.27&\cellgray5.23 \greendown&6.26&\cellgray5.24 \greendown&5.37&\cellgray5.24 \greendown&2.83&\cellgray3.40 \redup\\
			& DenseNet-121&20.99&\cellgray3.53(1.4) \greendown&5.63&\cellgray4.01 \greendown&5.61&\cellgray4.03 \greendown&5.54&\cellgray4.11 \greendown&4.84&\cellgray4.11 \greendown&3.16&\cellgray1.64 \greendown\\
			\midrule
			\multirow{4}{*}{ImageNet}& ResNet-50&3.70&\cellgray1.64(2.06) \greendown&2.08&\cellgray1.62 \greendown&2.07&\cellgray1.62 \greendown&2.07&\cellgray1.62 \greendown&3.05&\cellgray1.62 \greendown&1.32&\cellgray0.96 \greendown\\
            & DenseNet-121&6.66&\cellgray3.01(3.45) \greendown&1.65&\cellgray1.18 \greendown&1.66&\cellgray1.19 \greendown&1.64&\cellgray1.22 \greendown&2.25&\cellgray1.22 \greendown&2.61&\cellgray0.54 \greendown\\
            & Wide-Resnet-50&5.35&\cellgray2.58(3.05) \greendown&3.01&\cellgray2.35 \greendown&3.01&\cellgray2.35 \greendown&2.99&\cellgray2.34 \greendown&4.20&\cellgray2.34 \greendown&3.05&\cellgray0.92 \greendown\\
            & MobileNet-V2&2.71&\cellgray1.25(1.73) \greendown&1.88&\cellgray1.42 \greendown&1.87&\cellgray1.43 \greendown&1.93&\cellgray1.46 \greendown&2.21&\cellgray1.46 \greendown&1.80&\cellgray0.43 \greendown\\
			\bottomrule
		\end{tabular}%
	}
	\caption{\textbf{Adaptive ECE$\downarrow$ before and after after feature clipping.}\quad Adaptive ECE is measured as a percentage, with lower values indicating better calibration. Adaptive ECE is evaluated for different post hoc calibration methods, both before (base) and after (+ours) feature clipping. The results are calculated with number of bins set as 15. The optimal $c$ is determined on the validation set, included in brackets.}
	\label{table:adaptive ece compare with posthoc methods}
\end{table*}
\section{Full feature distribution}
Similar to the randomly selected feature units, we plot feature values for all 2,048 feature units, which show a similar pattern. The HCE samples exhibit a higher mean value in most units, with certain units displaying abnormally high mean values. We show the results in Fig.~\ref{fig:full_feature_distribution}.

\section{Calibration Performance on Adaptive ECE}
In addition to ECE, we evaluate different post-hoc calibration methods on Adaptive ECE, as shown in Table~\ref{table:adaptive ece compare with posthoc methods}. Similar to ECE, FC achieves consistent improvements across various methods. Furthermore, FC alone can outperform most of the previous methods.

\section{Train recipe for train-time calibration methods}
\label{training recipe}
All our experiments are run on single RTX3090 GPU. For training networks on CIFAR-10 and CIFAR-100, we using the pretrained weight provided by ~\citep{mukhoti2020calibrating}. We cite the training recipe provided by ~\citet{mukhoti2020calibrating} as following. We use SGD with a momentum of 0.9 as our optimiser, and train the networks for 350 epochs, with a learning rate of 0.1 for the first 150 epochs, 0.01 for the next 100 epochs, and 0.001 for the last 100 epochs. We use a training batch size of 128. Furthermore, we augment the training images by applying random crops and random horizontal flips.

\end{document}